%% file: main.tex
\documentclass[10pt,twocolumn,letterpaper]{article}

\makeatletter
\@namedef{ver@eso-pic.sty}{}
\makeatother
\let\LenToUnit\relax

\usepackage{iccv}
\usepackage{float}
\usepackage{pdfpages}
\usepackage[absolute,overlay]{textpos}
\usepackage{placeins}
\usepackage{amsmath}
\usepackage{amssymb}
\usepackage{multirow, booktabs}
\usepackage{makecell}
\usepackage{graphicx}
\usepackage{booktabs}
\usepackage{multirow}
\usepackage{array}
\usepackage{enumitem}
\usepackage{titling}

\definecolor{iccvblue}{rgb}{0.21,0.49,0.74}

\usepackage{listings}
\usepackage{xcolor} 
\usepackage[utf8]{inputenc}
\DeclareUnicodeCharacter{2014}{---} 
\DeclareUnicodeCharacter{05BF}{}    

\usepackage[pagebackref,breaklinks,colorlinks,allcolors=iccvblue]{hyperref}

\def\paperID{12178}
\def\confName{ICCV}
\def\confYear{2025}

\title{BADAS: Context Aware Collision Prediction Using Real-World Dashcam Data}
\author{
Roni Goldshmidt\\
{\tt\small roni.goldshmidt@getnexar.com}
\and 
Hamish Scott\\
{\tt\small hamish.scott@getnexar.com}
\and 
Lorenzo Niccolini \\
{\tt\small lorenzo.niccolini@getnexar.com}
\and 
Shizhan Zhu \\
{\tt\small shizhan.zhu@getnexar.com}
\and 
Daniel Moura \\
{\tt\small daniel.moura@getnexar.com}
\and 
Orly Zvitia \\
{\tt\small orly.zvitia@getnexar.com}
}

\begin{document}
\maketitle
\pagenumbering{gobble}

\input{sec/01_abstract}

\input{sec/02_introduction}
\input{sec/03_related_work}
\input{sec/04_methodology}
\input{sec/05_experiments}
\input{sec/06_discussion_conclusion}

{\small
\bibliographystyle{IEEEtran}
\bibliography{main}
}
\end{document}

%% file: sec/01_abstract.tex
\begin{abstract}
Existing collision prediction methods often fail to distinguish between ego-vehicle threats and random accidents non-involving ego-vehicle, leading to excessive false alerts in real-world deployment. We present BADAS, a family of collision prediction models trained on Nexar's real-world dashcam collision dataset—the first benchmark designed explicitly for ego-centric evaluation. We re-annotate major benchmarks to identify ego involvement, add consensus alert-time labels, and synthesize negatives where needed, enabling fair AP/AUC and temporal evaluation. BADAS uses a V-JEPA2 backbone trained end-to-end and comes in two variants: BADAS-Open (trained on our 1.5k public videos) and BADAS1.0 (trained on 40k proprietary videos). Across DAD, DADA-2000, DoTA, and Nexar, BADAS achieves state-of-the-art AP/AUC and outperforms a forward-collision ADAS baseline while producing more realistic time-to-accident estimates. We release our BADAS-Open model weights and code, along with re-annotations of all evaluation datasets to promote ego-centric collision prediction research.
\end{abstract}

%% file: sec/02_introduction.tex
\section{Introduction}
\label{sec:introduction}

\begin{figure}[!htb]
    \centering  
    \includegraphics[width=\columnwidth]{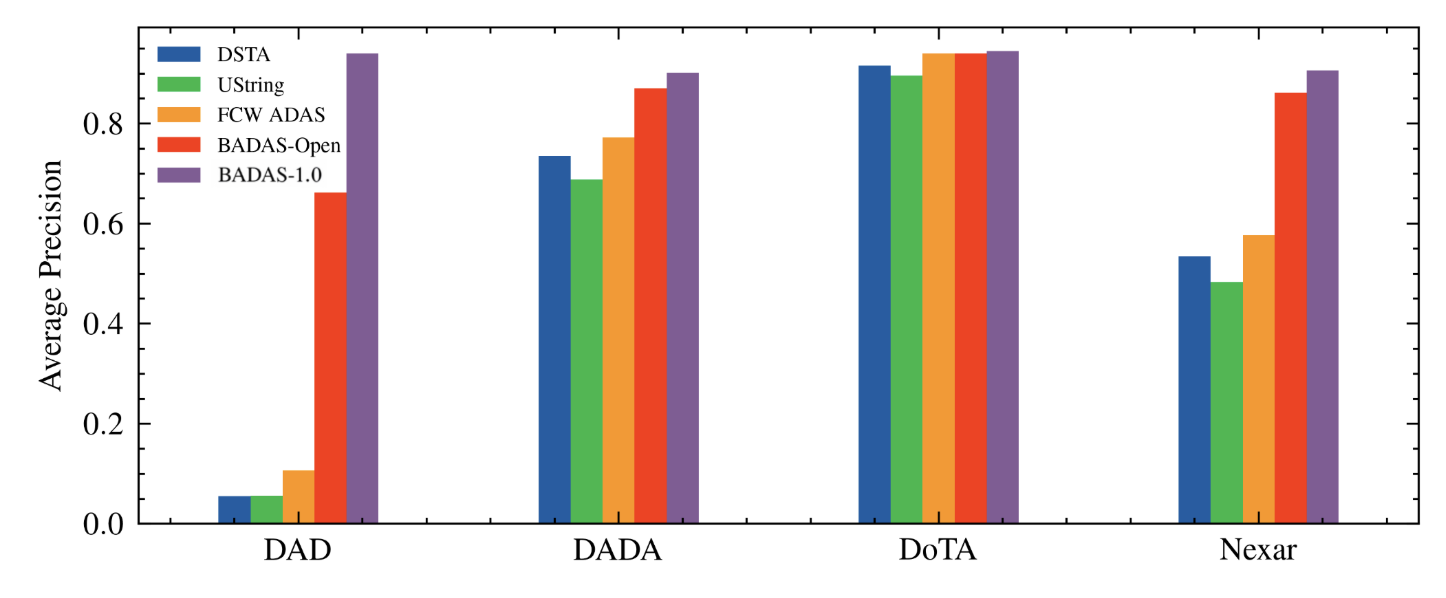}
    \caption{BADAS achieves state-of-the-art performance on collision prediction benchmarks. Our approach substantially outperforms existing academic methods and commercial vision-based forward collision warning systems by leveraging ego-centric real-world data and modern video foundation models when evaluated on ego-vehicle involved collisions.}
    \label{fig:performance_bar}
\end{figure}

Collision prediction is fundamental to Advanced Driver Assistance Systems (ADAS) and autonomous vehicles, yet current approaches fail to meet real-world deployment requirements. Despite decades of research, existing methods struggle with excessive false alarms and miss critical ego-vehicle threats. We present BADAS (V-JEPA2 ~\cite{assran2025vjepa} Based Advanced Driver Assistance System), a new approach that achieves state-of-the-art performance by combining modern video foundation models with high-quality, ego-centric real-world driving data. As shown in Figure~\ref{fig:performance_bar}, BADAS significantly outperforms both academic methods and commercial ADAS systems across major benchmarks, demonstrating the power of aligning training data with actual deployment scenarios.

The core challenge in collision prediction lies not in the algorithms themselves, but in the fundamental misalignment between training data and actual driving scenarios. More critically, existing real-world datasets like DAD~\cite{chan2017anticipating}, DoTA~\cite{yao2022dota}, and DADA-2000~\cite{fang2021dada} suffer from a conceptual flaw: they treat all visible accidents equally, training models to detect any collision within the camera's field of view. This approach generates excessive false alarms in deployment—a vehicle accident two lanes over triggers the same alert as an imminent frontal collision threatening the ego vehicle.

\begin{figure}[!htb]
    \centering  
    \includegraphics[width=\columnwidth]{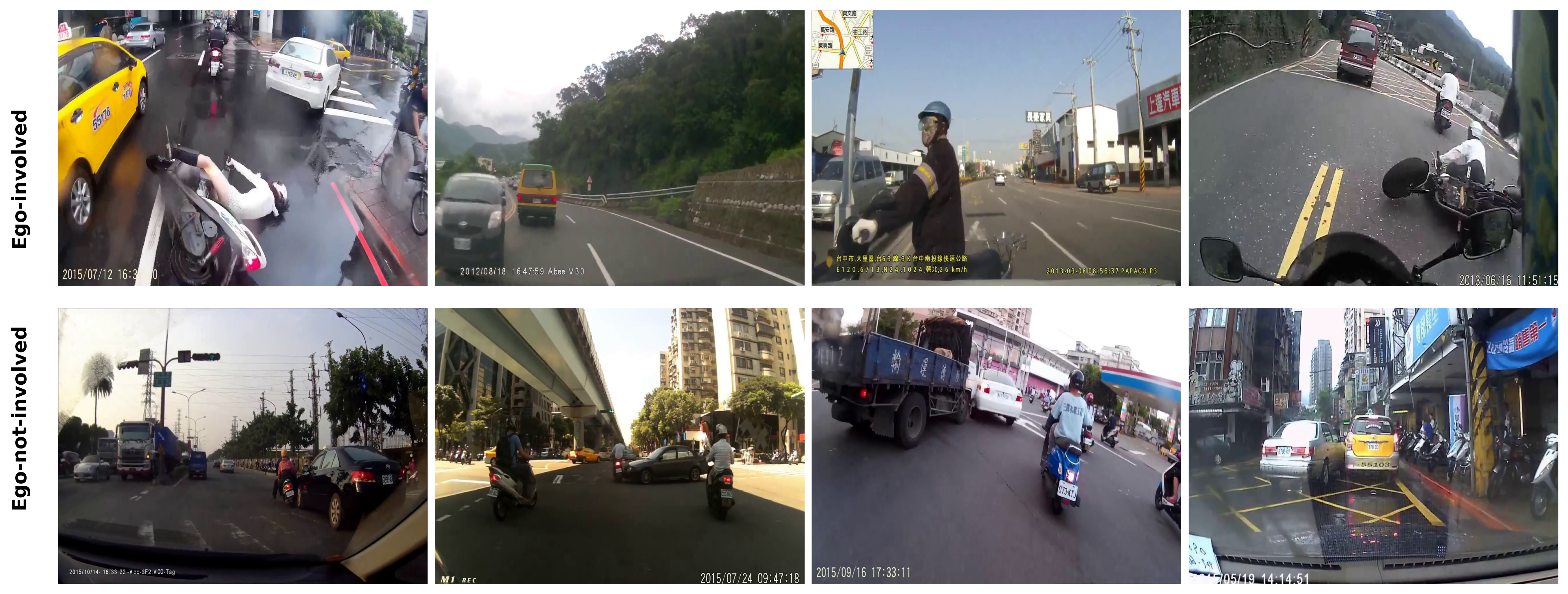}
    \caption{The critical distinction between ego-centric and general collision prediction. Top row: Ego-vehicle involved events requiring immediate driver intervention. Bottom row: Non-ego accidents that are visually prominent but irrelevant to ego-vehicle safety. BADAS focuses exclusively on ego-relevant scenarios to minimize false alarms while maintaining high sensitivity to actual threats. Examples from the DAD dataset~\cite{chan2017anticipating}.}
    \label{fig:ego_centric}
\end{figure}

Figure~\ref{fig:ego_centric} illustrates this critical distinction between ego-centric and general collision prediction. While visually salient accidents in adjacent lanes may capture attention, they are irrelevant to the ego vehicle's safety and should not trigger warnings. Our systematic re-annotation of existing benchmarks reveals the severity of this problem as shown in Table~\ref{tab:dataset_stats_filtered}. 

\begin{table}[!htb]
\centering
\caption[Re-annotation results of collision datasets]{%
Re-annotation results of major collision-prediction datasets.
The high percentage of non-ego accidents (40--92\%) suggests that these datasets should be used with caution when developing or evaluating ADAS-related methods.
\textit{Pos-Ego}: ego-involved accidents.
\textit{Pos-Not Ego}: non-ego accidents.
\textit{Less-2s}: samples filtered out due to insufficient anticipation time---specifically, cases where the collision occurs within 2 seconds from the beginning of the video. Our model requires a full 2 seconds prediction horizon; thus, such cases are invalid for anticipation.
\textit{Negative}: normal driving videos.
\textit{\%Not Ego}: percentage of non-ego accidents among retained positive samples.}
\label{tab:dataset_stats_filtered}
\resizebox{\columnwidth}{!}{%
\begin{tabular}{lccccc}
\toprule
Dataset & \#Pos-Ego & \#Pos-Not Ego & \#Less-2s & \#Negative & \% Pos-Not Ego \\
\midrule
DADA-2000~\cite{fang2021dada} & 75 & 51 & 2 & 0 & 40.5 \\
DAD~\cite{chan2017anticipating} & 13 & 150 & 2 & 301 & 92.0 \\
DoTA~\cite{yao2022dota} & 327 & 255 & 16 & 0 & 43.8 \\
Nexar~\cite{nexar2025challenge} & 672 & 0 & 0 & 672 & 0.0 \\
\bottomrule
\end{tabular}%
}
\end{table}

Beyond the ego-centric issue, existing datasets exhibit additional limitations, posing challenges for effective ADAS development. DoTA and DADA-2000 contain only positive examples, making it impossible to evaluate false positive rates—a crucial metric for user acceptance. They also lack consistency in defining alert timing: DAD bases alerts on abnormal behavior onset, DoTA uses subjective "inevitability" judgments, and DADA-2000 triggers alerts upon 3rd party vehicle appearance regardless of actual risk. 

The recently introduced Nexar Dashcam Collision Prediction Dataset~\cite{nexar2025challenge, moura2025nexar} addresses these fundamental problems through a paradigm shift in data collection. Unlike existing datasets, Nexar comprises 1.5k real-world dashcam videos from actual drivers experiencing genuine collisions and near-misses. This dataset provides three key advantages: (1) exclusive focus on ego-vehicle involved events, eliminating irrelevant accidents, (2) inclusion of near-collision events where accidents are avoided through emergency maneuvers—capturing the much more common successfully-resolved dangerous situations that provide rich training signals, and (3) standardized, consensus-based alert timing annotations from 10 human annotators that enable consistent temporal evaluation. The real-world nature of this data is crucial: it captures the true distribution of driving scenarios, including rare events that synthetic datasets cannot replicate and that drivers actually encounter on the road.

Our approach leverages recent advances in video foundation models, particularly V-JEPA2 ~\cite{assran2025vjepa}, which has shown strong capabilities in understanding temporal dynamics and visual patterns. These architectures excel when trained on data that connects visual patterns to concrete outcomes—precisely what ego-centric collision and near-collision events provide. By fine-tuning V-JEPA2 on the Nexar high-quality ego-relevant dataset, we harness its temporal reasoning capabilities for collision prediction. The combination of modern architectures with appropriate real-world data yields the significant performance gains shown in Figure~\ref{fig:performance_bar}, substantially surpassing both task-specific architectures and commercial ADAS systems.

The contributions of this paper are as follows:
\begin{enumerate}
    \item \textbf{Ego-centric problem reformulation}: We redefine collision prediction as an ego-centric task and systematically re-annotate major benchmarks (DAD, DoTA, DADA-2000) to identify ego-vehicle involvement, revealing that 40-92\% of their accidents are irrelevant for ADAS applications. These annotations are publicly released to benefit the community.
    
    \item \textbf{Standardized temporal evaluation}: We establish a coherent definition of alert timing through 10-annotator consensus and contribute precise temporal annotations for all test sets, addressing the inconsistent and subjective definitions across existing datasets. Our results in \ref{fig:tta_boxplot} emphasize that longer estimated time to accident may represent early anticipations rather than a better actual prediction.
    
    \item \textbf{State-of-the-art performance}: Our models surpass existing methods by fine-tuning V-JEPA2 ~\cite{assran2025vjepa} on high-quality, ego-centric real-world data, demonstrating that combining advanced architectures with appropriate data significantly outperforms both academic methods and commercial vision-based forward collision warning systems.
    
    \item \textbf{Real-world data importance}: We demonstrate that practical collision prediction requires training in genuine traffic situations rather than synthetic or controlled environments. A preliminary analysis (Figure 9a) already reveals the natural long-tailed structure of real-world collisions, underscoring the value of diverse large-scale data.
\end{enumerate}

The rest of this paper is organized as follows: Section 2 reviews related work in collision prediction datasets and methods and video understanding. Section 3 presents our methodology including model architecture and training protocols. Section 4 describes our experimental setup and re-annotation process. Section 5 presents a comprehensive analysis and results. We conclude in Section 6 and discuss potential directions for future research in Section 7.

%% file: sec/03_related_work.tex
\section{Related Work}

\subsection{Traffic Accident Prediction Methods}
Recent advances in traffic accident prediction have produced various approaches, though we focus our comparison on methods with open-source implementations that enable reproducible evaluation on our ego-centric test sets.

UString~\cite{suzuki2018anticipating} uses RNN-based architectures with adaptive loss functions that emphasize frames near collision events, dynamically adjusting loss contribution based on temporal proximity to accidents. DSTA~\cite{karim2022dynamic} employs transformer-based architectures with dynamic spatial-temporal attention mechanisms, allowing the model to focus on relevant spatial regions while tracking their temporal evolution. Both methods provide open-source implementations, facilitating direct comparison.

While other approaches using graph neural networks~\cite{malawade2022spatiotemporal}, reinforcement learning~\cite{bao2021drive}, and multi-modal fusion~\cite{fang2024abductive} exist in the literature, the lack of publicly available implementations prevents fair comparison on our ego-centric datasets. Therefore, our experimental comparison focuses on UString and DSTA as the current open-source state-of-the-art.

\subsection{Collision Prediction Datasets and Temporal Annotations}
Existing datasets differ significantly in their temporal annotation approaches. DAD~\cite{chan2017anticipating} contains 1,750 videos and defines accidents based on abnormal behavior onset rather than impact moments. DoTA~\cite{yao2022dota} provides 4,990 videos with subjective annotations marking when accidents appear "inevitable," resulting in inconsistent temporal boundaries. DADA-2000~\cite{fang2021dada} offers 1,962 videos annotating from vehicle appearance to collision, which may mark accidents as predictable too early when vehicle presence alone doesn't indicate danger.

The Nexar dataset~\cite{moura2025nexar} addresses these limitations through consensus-based alert times from multiple annotators, establishing both the earliest moment of recognizable danger ($t_{\text{alert}}$) and precise collision timing ($t_{\text{collision}}$), enabling evaluation of both prediction accuracy and temporal appropriateness.

\subsection{Foundation Models for Video Understanding}
Video foundation models have evolved from temporal extensions of image architectures to sophisticated self-supervised approaches. SlowFast networks~\cite{feichtenhofer2019slowfast} process videos at multiple temporal resolutions, while Video Swin Transformers~\cite{liu2022video} extend window-based attention temporally. 

Self-supervised methods like VideoMAE~\cite{tong2022videomae} reconstruct masked spatial-temporal patches, while V-JEPA~\cite{bardes2024revisiting} predicts abstract representations rather than raw pixels. V-JEPA2 ~\cite{assran2025vjepa} further improves masking strategies and training objectives for temporal understanding, making it particularly suited for anticipatory tasks. BADAS is the first application of V-JEPA2 for collision prediction, demonstrating that fine-tuning on ego-centric data substantially outperforms task-specific architectures.

\subsection{Industrial ADAS Systems}
Current ADAS Forward Collision Warning (FCW) systems typically combine radar and camera sensors with physics-based models to calculate time-to-collision and trigger alerts~\cite{EuroNCAP2020}. For comparison purposes, we focus on vision-only implementations that process monocular dashcam inputs. The open-source FCW system~\cite{VehicleCVADAS} uses YOLO~\cite{Khanam2024YOLOv11} for object detection and~\cite{qin2022ultrav2} for lane detection, raising alerts when detected objects fall within distance thresholds. This provides a baseline representing current deployed vision-based ADAS technology.

%% file: sec/04_methodology.tex
\section{Methodology}
\label{sec:methodology}

\subsection{Problem Formulation}

Building on the Nexar Challenge~\cite{nexar2025challenge}, we reformulate collision prediction as an ego-centric task. Given a dashcam video sequence $\mathcal{V} = \{f_1, ..., f_T\}$, we predict at each timestep $t$ the probability $p_t = P(\text{ego-collision}|f_{1:t})$ that the ego vehicle will be involved in a collision within time horizon $\tau$.

\textbf{Ego-vehicle incidents:} Scenarios where the ego vehicle experiences collision or executes emergency maneuvers, including direct collisions, near-misses requiring evasive action, and situations prevented only by emergency intervention.

\textbf{Non-ego incidents:} Accidents visible but irrelevant to ego-vehicle safety, including adjacent lane collisions and distant events.

Near-collisions (emergency maneuvers that prevented accidents) are treated as positive training examples, providing rich supervisory signals from successfully-resolved dangerous situations.

\subsection{BADAS Architecture}

BADAS uses V-JEPA2 ~\cite{assran2025vjepa} backbone with patch aggregation and classification head:

\textbf{V-JEPA2 Backbone:} Encoder patchifies videos using $2 \times 16 \times 16$ tubelets and passes tokens through ViT-L transformer. We process 16 frames of size $256 \times 256$, yielding 2048 latent patches with dimension $D=1024$.

\textbf{Patch Aggregation:} We aggregate patches using attentive probe~\cite{psomas2025attentionpleaserevisitingattentive}. Given patches $X \in \mathbb{R}^{P \times D}$, learned queries $Q \in \mathbb{R}^{M \times D}$ compute attention scores $A = \text{softmax}(QX^\top/\sqrt{D})$. These aggregate patches via weight matrix $W \in \mathbb{R}^{D\times d}$ to produce features $AXW \in \mathbb{R}^{M \times d}$, concatenated to final vector of size $Md$ (with $M=12$, $d=64$).

\textbf{Prediction Head:} Three-layer MLP with GELU activation, layer normalization, and 0.1 dropout probability maps aggregated features to collision probability. Hidden dimension: 768.

\subsection{Training Protocol}

We fine-tune V-JEPA2 end-to-end using AdamW optimizer with learning rate $\eta = 1 \times 10^{-5}$, weight decay $1 \times 10^{-4}$, and cosine annealing schedule. Binary cross-entropy loss with mixed precision training and gradient clipping (norm 5.0) ensures stability. Early stopping on validation set prevents overfitting.

\subsection{Dataset Re-annotation Protocol}

We systematically re-annotated DAD, DoTA, and DADA-2000 for ego-centric evaluation:

\textbf{Ego-Involvement Classification:} We manually reviewed all accidents, categorizing them as ego-involved - recording vehicle is directly involved in the accident, or non-ego involved - accidents in adjacent lanes, perpendicular traffic or visible but non-threatening.

\textbf{Human Alert Time:} 10 annotators with defensive driving certification marked when they would initiate defensive action. Consensus time (median) represents realistic human response timing.

\begin{figure}[H]
\centering
\includegraphics[width=\columnwidth]{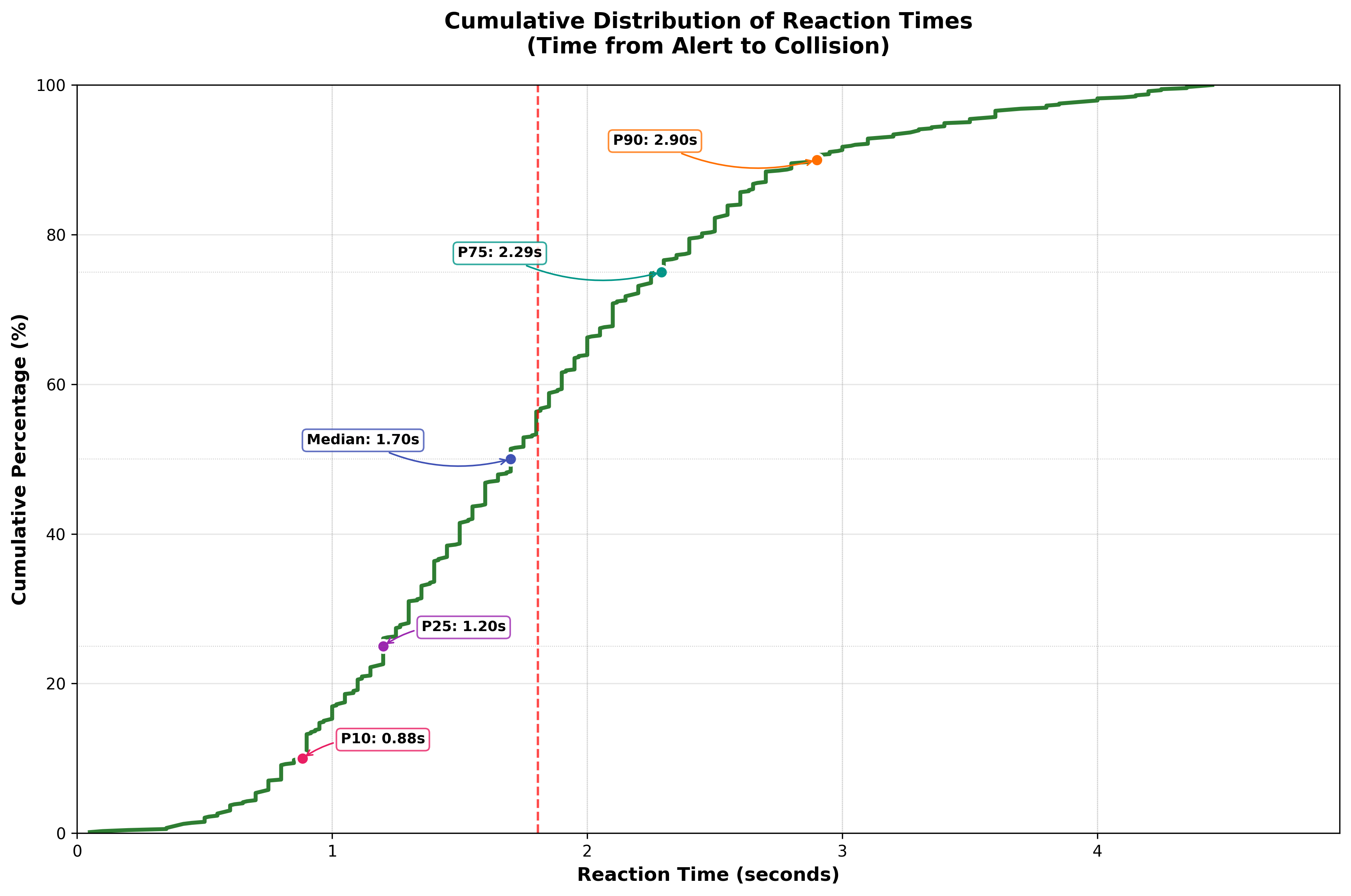}
\caption{Cumulative distribution of human reaction times across 726 ego-involved collisions. Median: 1.70s, 90\% range: 0.70-3.47s before impact.}
\label{fig:reaction_time_cdf}
\end{figure}

Figure~\ref{fig:reaction_time_cdf} shows human reaction patterns across 726 events: median 1.70s, mean 1.81s (SD=0.82s), with 90\% of alerts between 0.70-3.47s before collision. Alerts beyond 95th percentile (3.47s) likely respond to normal variations rather than genuine threats.

\textbf{Synthetic Negatives:} For datasets lacking negative samples, we extract first 4 seconds from videos where human alert occurs 4.5s or later after video start, ensuring realistic driving without collision precursors.

\begin{table}[H]
\centering
\caption{Dataset composition after ego-centric re-annotation}
\label{tab:dataset_composition}
\footnotesize
\setlength{\tabcolsep}{4pt}
\begin{tabular}{@{}lcccc@{}}
\toprule
\textbf{Dataset} & \textbf{Real Neg.} & \textbf{Real Pos.} & \textbf{Synth. Neg.} & \textbf{Total} \\
\midrule
DADA-2000 & 0 & 75 & 38 & 113 \\
DAD & 301 & 13 & 0 & 314 \\
DoTA & 0 & 327 & 40 & 367 \\
\bottomrule
\end{tabular}
\end{table}

Our consensus-based protocol provides unified temporal reference across datasets, addressing the existing inconsistencies and enabling fair comparison. The re-annotations are publicly available \footnote{\url{https://github.com/getnexar/BADAS-Open/tree/main}}.

\subsection{Model Variants}

\textbf{BADAS-Open:} Trained solely on Nexar's public dataset (1,500 videos) with balanced collision/normal driving representation. Achieves state-of-the-art on major benchmarks. Model and code are publicly available \footnote{\url{https://huggingface.co/nexar-ai/BADAS-Open
}} 
under the APSv2 license.

\textbf{BADAS1.0:} Commercial variant trained on 40k proprietary sequences (20k collisions/near-misses, 20k normal driving). Uses identical architecture and training protocols, but 25× more data, enabling data scaling effect evaluation and superior edge-case handling.

%% file: sec/05_experiments.tex
\section{Experiments}
\label{sec:experiments}

\subsection{Experimental Setup}

We evaluate BADAS against state-of-the-art collision prediction methods and industrial ADAS systems. Our baseline methods include UString~\cite{suzuki2018anticipating}, DSTA~\cite{karim2022dynamic} and FCW ADAS~\cite{VehicleCVADAS}, a vision-only forward collision warning system using YOLOv11~\cite{Khanam2024YOLOv11} for object detection and TuSimple ResNet-34~\cite{qin2022ultrav2} for lane detection.

Evaluation is performed on three benchmarks re-annotated for ego-centric assessment: DAD~\cite{chan2017anticipating}, DADA~\cite{fang2021dada}, DoTA~\cite{yao2022dota}, and the Nexar test set. For datasets originally containing only positive samples (DADA, DoTA), we generate synthetic negatives as described in Section~\ref{sec:methodology}. Performance is measured using Average Precision (AP) for ranking quality, Area Under Curve (AUC) for discrimination capability, and mean Time-To-Accident (mTTA) for temporal accuracy.

\section{Results}

\subsection{Quantitative Performance}

Table~\ref{tab:main_results} presents a comprehensive summary of our evaluation results. BADAS models achieve substantial improvements over existing methods, with BADAS1.0 reaching 0.94 AP on DAD compared to 0.06 for baseline methods. The performance gap is particularly striking on the Nexar dataset—the largest and most comprehensive benchmark—where BADAS-Open achieves 0.86 AP versus 0.48-0.53 for academic baselines and 0.58 for the industrial FCW system.

\begin{table}[t]
\centering
\caption{Ego-centric collision prediction performance across benchmarks.}
\label{tab:main_results}
\resizebox{\columnwidth}{!}{
    \begin{tabular}{l|ccc|ccc|ccc|ccc}
    \toprule
     & \multicolumn{3}{c|}{\textbf{DAD (n=116)}} & \multicolumn{3}{c|}{\textbf{DADA (n=113)}} & \multicolumn{3}{c|}{\textbf{DoTA (n=367)}} & \multicolumn{3}{c}{\textbf{Nexar (n=1344)}} \\
    \cmidrule{2-13}
    \textbf{Method} & AP & AUC & mTTA & AP & AUC & mTTA & AP & AUC & mTTA & AP & AUC & mTTA \\
    \midrule
    DSTA & 0.06 & 0.59 & 2.9 & 0.74 & 0.60 & 6.5 & 0.92 & 0.55 & 4.8 & 0.53 & 0.54 & 9.8 \\
    UString & 0.06 & 0.61 & 2.4 & 0.69 & 0.57 & 5.3 & 0.90 & 0.54 & 4.3 & 0.48 & 0.48 & 9.1 \\
    FCW ADAS & 0.11 & 0.69 & 2.0 & 0.77 & 0.78 & 3.8 & 0.94 & 0.69 & 3.0 & 0.58 & 0.64 & 8.7 \\
    \hline
    BADAS-Open & 0.66 & 0.87 & 2.7 & 0.87 & 0.77 & 4.3 & 0.94 & 0.70 & 4.0 & 0.86 & 0.88 & 4.9 \\
    \textbf{BADAS1.0} & \textbf{0.94} & \textbf{0.99} & 2.7 & \textbf{0.90} & \textbf{0.87} & 4.6 & \textbf{0.95} & \textbf{0.72} & 4.0 & \textbf{0.91} & \textbf{0.91} & 3.9 \\
    \bottomrule
    \end{tabular}
}
\end{table}

Notably, BADAS models maintain consistent performance across diverse datasets, indicating robust generalization. The mTTA values reveal a critical distinction: baseline methods report physically implausible predictions (9-10 seconds before collision on Nexar), while BADAS maintains more realistic 3-5 second windows better aligned with human prediction capabilities. The industrial FCW system, despite using state-of-the-art detection models, underperformed due to its rule-based logic generating excessive false positives in dense traffic—highlighting the advantages of end-to-end learning from real collision data.

\subsection{Qualitative Analysis}

\begin{figure}[t]
    \centering
    \includegraphics[width=\columnwidth]{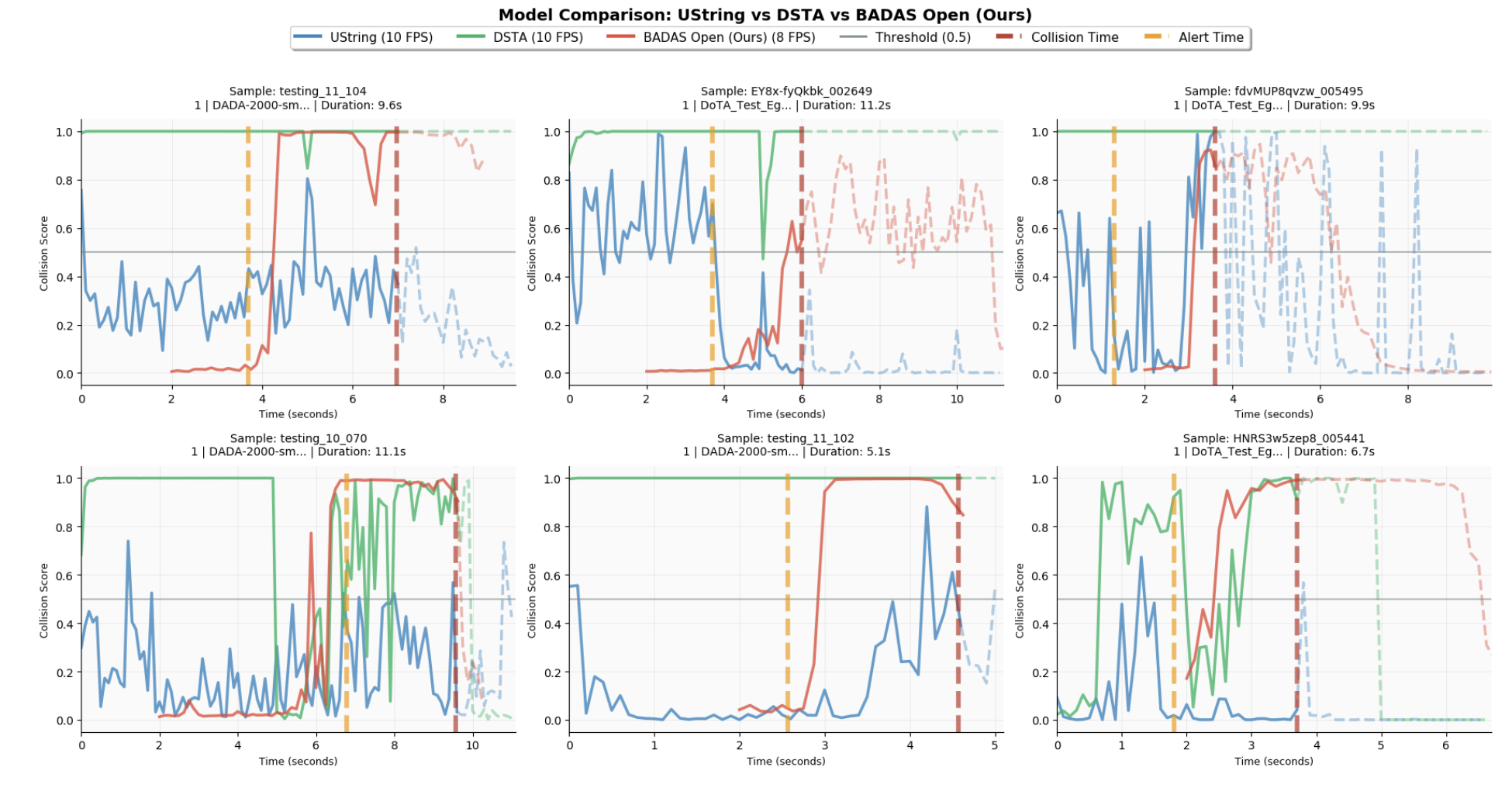}
    \caption{Collision predictions scores over time for 6 samples: UString (blue), DSTA (green) and BADAS-Open (red). Vertical dashed lines indicate collision time, horizontal line shows detection threshold.}
    \label{fig:visual_comparison} 
\end{figure}

Figure~\ref{fig:visual_comparison} illustrates prediction scores over time for representative test samples. BADAS-Open (red) demonstrates stable, confident predictions that rise sharply as collisions approach, while baseline methods exhibit erratic patterns with premature or inconsistent alerting. This stability translates to practical deployment benefits leading to fewer false alarms and more reliable threat assessment.

\begin{figure*}[t]
    \centering
    \includegraphics[width=\textwidth]{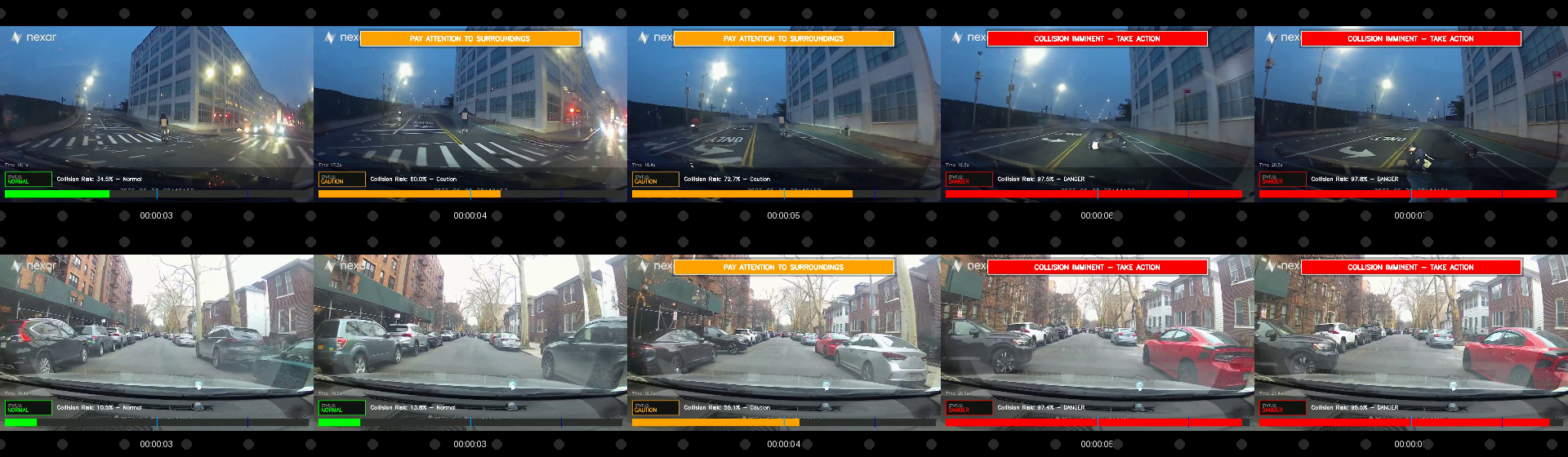}
    \caption{Frame-by-frame comparison on Nexar test footage. Top: urban intersection collision. Bottom: residential street scenario. Alert indicators show prediction confidence (green: safe, orange: caution, red: imminent collision). BADAS variants demonstrate accurate ego-centric threat assessment.}
    \label{fig:video_frames_comparison}
\end{figure*}

Frame-by-frame analysis in Figure~\ref{fig:video_frames_comparison} further demonstrates that in urban intersection and residential scenarios, BADAS variants correctly identify ego-relevant threats while maintaining low confidence during safe driving periods. 

\subsection{Data Scaling Effects}

\begin{figure}[t]
    \centering
    \includegraphics[width=\columnwidth]{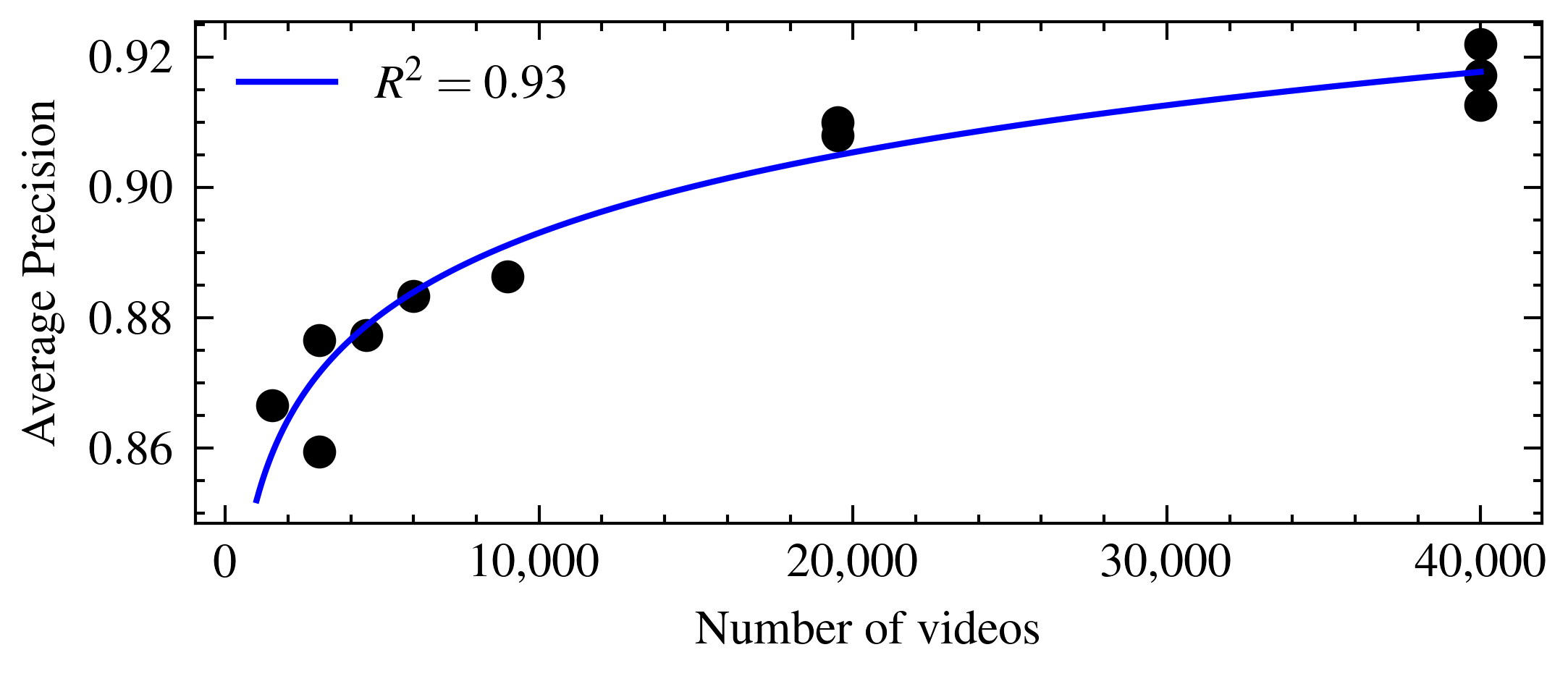}
    \caption{Validation AP versus training data size, showing logarithmic performance improvement with data scaling.}
    \label{fig:learning_curves}
\end{figure}

Figure~\ref{fig:learning_curves} illustrates the effect of training data scale, showing a logarithmic improvement in Nexar validation AP as the dataset size increases from 1.5k to 40k videos. This consistent scaling trend supports our hypothesis that leveraging large-scale real-world data continues to yield performance gains, even at the largest scales tested.

\subsection{Ablation Studies}

We investigate key design choices through systematic ablations. Table~\ref{tab:ablation_label_window} examines label window selection, showing that $T=1.5$ seconds optimally balances precision (0.785) and recall (0.939). Shorter windows miss developing collision patterns, while longer windows include too many non-indicative frames.

\begin{table}[t]
    \centering
    \caption{Performance for different label windowing strategies.}
    \label{tab:ablation_label_window}
    \begin{tabular}{c|ccccc}
    \toprule
    Window & Precision & Recall & AP & AUC & mTTA \\
    \midrule
    1.0s & \textbf{0.792} & 0.896 & 0.930 & 0.930 & 3.112 \\
    1.5s & 0.785 & 0.939 & \textbf{0.931} & \textbf{0.935} & 3.880 \\
    2.0s & 0.741 & \textbf{0.964} & 0.925 & 0.931 & \textbf{4.646} \\
    \bottomrule
    \end{tabular}
\end{table}

Data augmentation and oversampling strategies (Table~\ref{tab:ablation_data_sampling}) show significant benefits, with the combination of 2× oversampling for positive samples and augmentations improving AP from 0.922 to 0.939. The augmentations (Figure~\ref{fig:augmentations}) simulate diverse weather and lighting conditions, improving robustness to real-world variations.

\begin{table}[t]
    \centering
    \caption{Impact of sampling strategies and augmentations.}
    \label{tab:ablation_data_sampling}
    \begin{tabular}{c|c|cc}
    \toprule
     Oversampling Rate & Augmentations & AP & AUC \\
    \midrule
    1× & No & 0.922 & 0.927 \\
    2× & No & 0.924 & 0.922 \\
    1× & Yes & 0.931 & 0.935 \\
    2× & Yes & \textbf{0.939} & \textbf{0.941} \\
    \bottomrule
    \end{tabular}
\end{table}

\begin{figure}[h]
\centering
\begin{lstlisting}[language=Python]
import albumentations as A

A.Compose([
    A.RandomBrightnessContrast(p=0.5),
    A.HueSaturationValue(p=0.5),
    A.GaussianBlur(blur_limit=(3, 7), p=0.3),
    A.CLAHE(p=0.3),
    A.RandomGamma(p=0.3),
    A.RGBShift(p=0.3),
    A.MotionBlur(blur_limit=7, p=0.3),
    A.RandomRain(p=0.2),
    A.RandomSnow(p=0.2),
    A.RandomShadow(p=0.2),
])
\end{lstlisting}
\caption{Data augmentations for weather and lighting robustness.}
\label{fig:augmentations}
\end{figure}

Architecture ablations (Table~\ref{tab:ablation_model_arch}) show the contribution of each component. We train models with the base V-JEPA2 backbone with: only a linear head, the attentive probe with a linear head, and the attentive probe with the MLP head. We test training with and without the base model frozen. Fine-tuning the V-JEPA2 backbone provides the largest gain (0.707→0.928 AP), while the attentive probe and MLP head add incremental improvements, reaching 0.939 AP with the complete architecture.

\begin{table}[t]
    \centering
    \caption{Architecture component contributions.}
    \label{tab:ablation_model_arch}
    \begin{tabular}{l|cc}
    \toprule
    Configuration & AP & AUC \\
    \midrule
    Frozen base & 0.707 & 0.744 \\
    Frozen base + probe & 0.782 & 0.811 \\
    Frozen base + probe + MLP & 0.771 & 0.809 \\
    Fine-tuned base & 0.928 & 0.935 \\
    Fine-tuned base + probe & 0.929 & 0.936 \\
    Fine-tuned base + probe + MLP & \textbf{0.939} & \textbf{0.941} \\
    \bottomrule
    \end{tabular}
\end{table}

\subsection{Temporal Accuracy Analysis}

\begin{figure}[h!]
    \centering
    \includegraphics[width=\columnwidth]{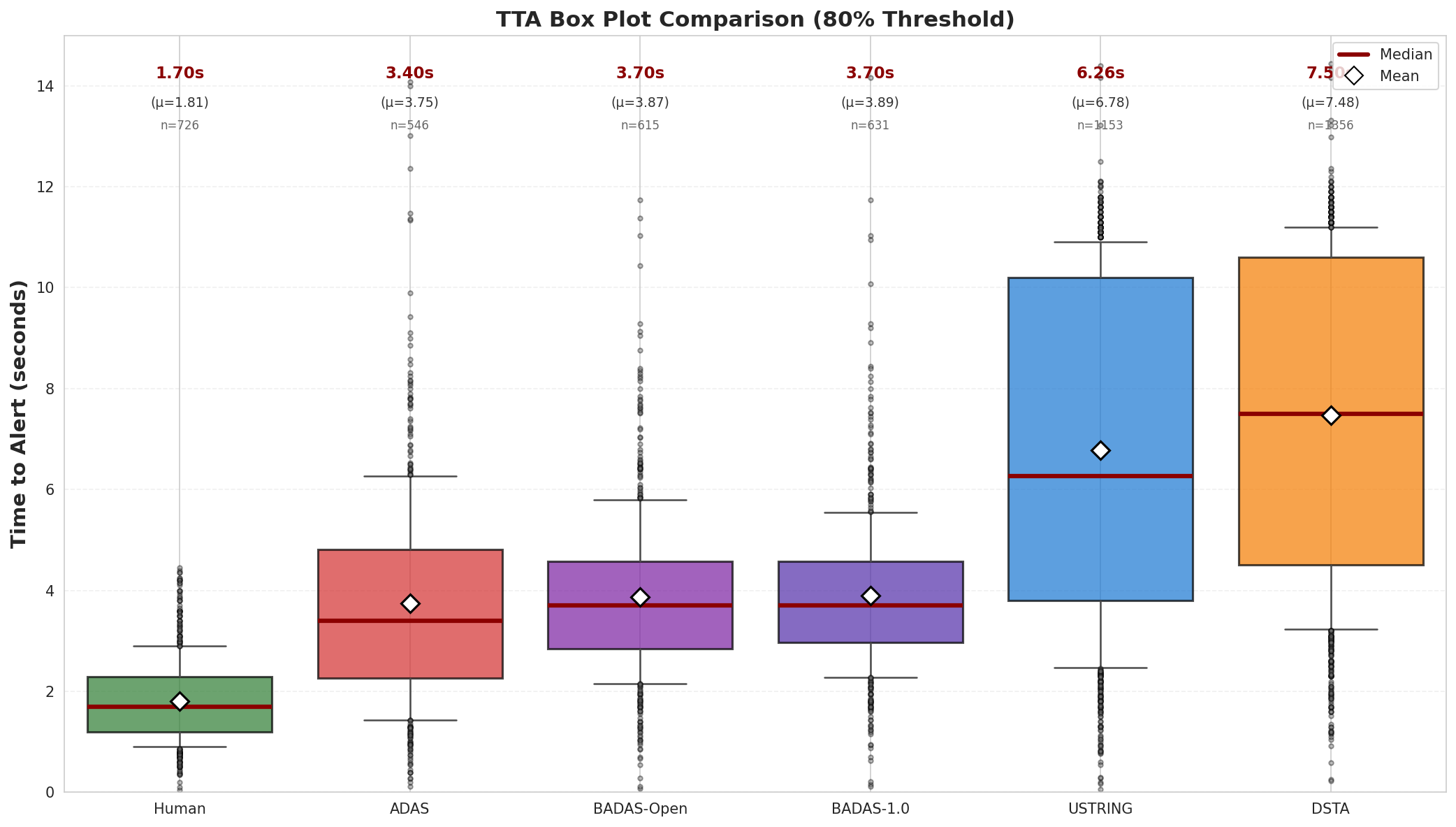}
    \caption{TTA distributions at 80\% confidence. Human consensus (green) shows expert annotations. BADAS-Open demonstrates realistic timing while baselines exhibit implausibly early predictions.}
    \label{fig:tta_boxplot}
\end{figure}

Figure~\ref{fig:tta_boxplot} compares Time-To-Accident distributions at 80\% confidence against human consensus annotations. BADAS-Open's median TTA of 3.0 seconds closely aligns with practical requirements—providing sufficient warning while avoiding premature alerts. In contrast, UString and DSTA show median TTAs of 6.2s and 7.5s respectively, with high variance extending to implausible 14-second predictions. This unrealistic early detection often manifests as false positives in deployment, posing challenges when integrating these methods into real-world ADAS.

\subsection{Long-Tail Performance}

Beyond common vehicle-to-vehicle interactions, real-world collision prediction must also recognize rare but safety-critical scenarios. To assess BADAS-Open robustness under such conditions, we measure recall across various third-party long-tail categories for confidence threshold of 0.85 relative to the recall observed on the dominant "vehicle" class.

\begin{figure}[h!]
  \centering
  \includegraphics[width=0.9\linewidth]{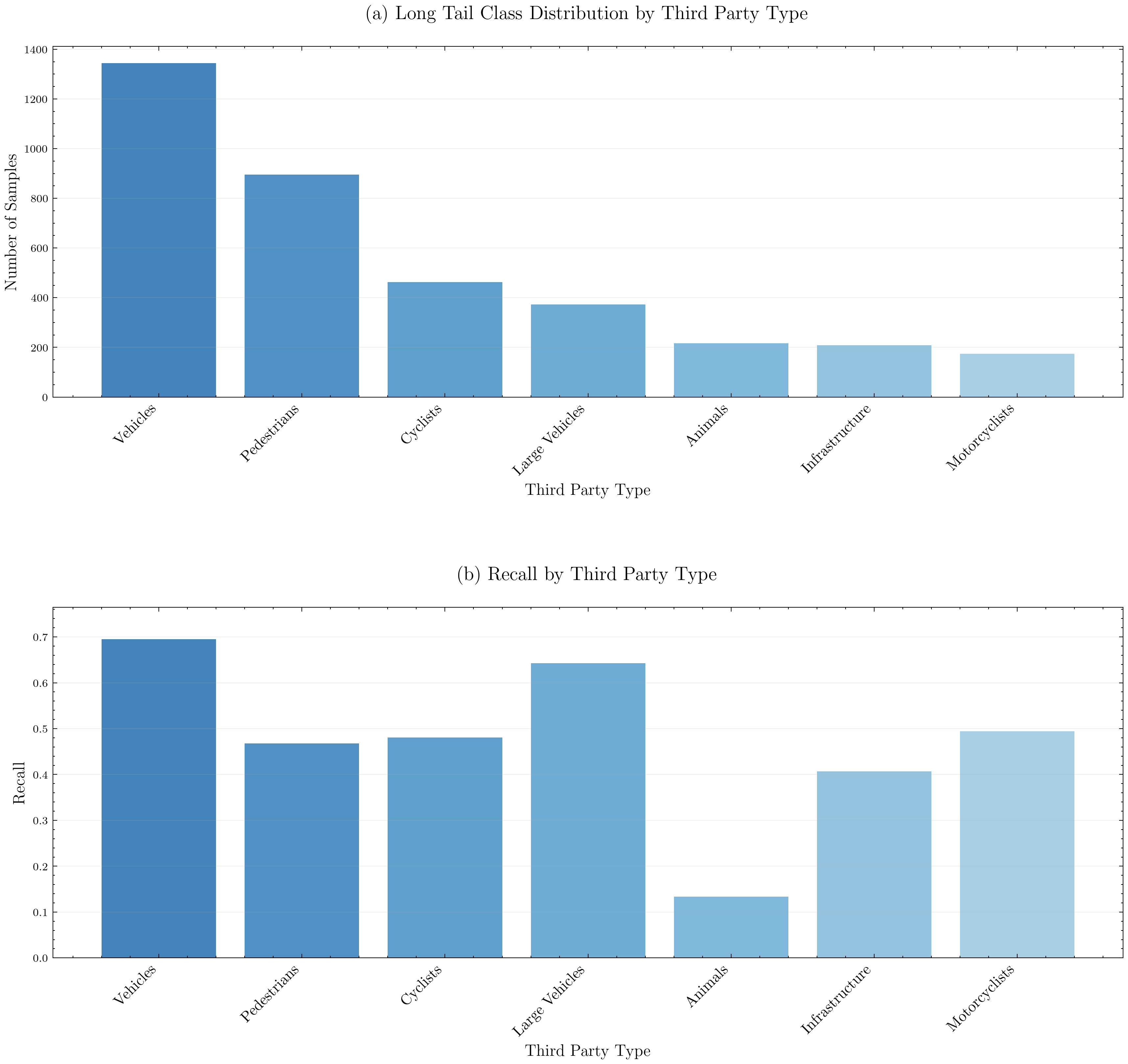}
\caption{(a) Distribution of third-party categories, representing the interacting agents. Static infrastructure include  buildings, fences, poles and traffic signs. The vehicle class corresponds to the Nexar \cite{nexar2025challenge} test set used throughout this work, while the remaining categories originate from additional annotated data. (b) BADAS-Open recall across these categories at a confidence threshold of 0.85}
  \label{fig:longtail_auc}
\end{figure}

As shown in Figure~\ref{fig:longtail_auc}, the model achieves the highest recall on vehicle-related events, with only a modest drop in larger vehicles (trucks and busses). Vulnerable Road Users (VRUs) categories - pedestrians, cyclists, motorcyclists - show a notable decline in recall, and performance decreases sharply further on animal-related incidents. 
\\ 
This degradation highlights the challenges of generalizing to rare, visually diverse, and low-frequency event types. While this behavior is expected given the absence of such cases in the training distribution of the BADAS-Open model, these results underscore the importance of explicitly evaluating the long-tail performance in collision prediction. 

%% file: sec/06_discussion_conclusion.tex

\section{Conclusion}
\label{sec:conclusion}

This work introduces BADAS, a new approach to collision prediction that focuses on ego-vehicle safety through ego-centric problem formulation. Building on insights from the Nexar Dashcam Collision Prediction Challenge, we demonstrate that focusing exclusively on ego-vehicle threats—rather than general accident detection—dramatically improves real-world performance.

Our systematic re-annotation of major benchmarks reveals fundamental issues with existing datasets: a significant portion of annotated accidents do not involve the ego-vehicle, leading models to learn patterns irrelevant to ego-vehicle safety. By filtering for ego-relevance and establishing human baseline reaction times, we create evaluation protocols that better reflect real-world deployment requirements. Our synthetic negative sampling method further improves the balance between positive and negative samples and relaxes the biased AP and AUC measurements. 

We further highlight the necessity of a coherent definition and annotation scheme for alert time, to serve as reference to the predicted mTTA values. Our findings show varying levels of early prediction in all methods. This is especially important for practical systems as these early predictions will be manifested as false alerts when deployed in real ADAS or AV frameworks.   

We present two model variants addressing different deployment needs: BADAS-Open, trained exclusively on 1.5k public Nexar videos, and BADAS1.0, leveraging 40k videos from Nexar's proprietary dataset. Both models achieve state-of-the-art performance when compared to leading research methods and FWC systems. The significant performance gain observed with increased data volume suggests that the potential of data scaling has not yet been fully saturated. The BADAS-Open model and code are released to the research community.

While our model outperforms existing state-of-the-art results, we also highlight the long-tail nature of collision and near-collision distributions, showing that BADAS-Open performance significantly deteriorates on minority classes.
This result is expected, as any model trained on an imbalanced dataset naturally focuses on majority classes (e.g., vehicle-to-vehicle accidents).
However, edge cases must also be taken into account — first by explicitly evaluating current model performance on them, and later by developing dedicated strategies to improve their prediction.

\section{Future Work}
While this study provides encouraging evidence for the effectiveness of context-aware architectures in collision prediction, several open challenges remain.

Future research directions include expanding the dataset to further enhance generalization, improving mean time-to-alert (mTTA) to reduce false alerts in real-world systems, and addressing long-tail categories to better evaluate and predict diverse and rare driving scenarios. 
Our model ability to recognize complex and risky situations even before collisions occurred as illustrated in Figure \ref{fig:video_frames_comparison} suggests the potential to extend collision prediction models beyond a binary formulation, toward a three-level taxonomy: \emph{normal, warning, and alert}. Such an approach could be particularly beneficial for autonomous driving systems, enabling adaptive decision-making based on momentary risk levels. We refer the readers to our project page for full length examples \footnote{\url{https://www.nexar-ai.com/badas}}. 

Ultimately, advancing reliable and context-aware collision prediction can contribute significantly to the broader goal of safer, more anticipatory driver assistance systems, and may play a key role in bridging the gap between current ADAS technologies and fully autonomous driving.